# A Cluster-Based Weighted Feature Similarity Moving Target Tracking Algorithm for Automotive FMCW Radar

Rongqian Chen
School of Information Science and Technology
South-west Jiaotong University
Chengdu, China
willchan@seas.upenn.edu

Yingquan Zou
School of Information Science and Technology
South-west Jiaotong University
Chengdu, China
zouyingq@swjtu.edu.cn

Anyong Gao
School of Information Science and Technology
South-west Jiaotong University
Chengdu, China
1306885376@qq.com

Leshi Chen
School of Computing and Artificial Intelligence
South-west Jiaotong University
Chengdu, China
chelsea_sushi@outlook.com

*Abstract*—We studied a target tracking algorithm based on millimeter-wave (MMW) radar in an autonomous driving environment. Aiming at the cluster matching in the target tracking stage, a new weighted feature similarity algorithm is proposed, which increases the matching rate of the same target in adjacent frames under strong environmental noise and multiple interference targets. For autonomous driving scenarios, we constructed a method that uses its motion parameters to extract and correct the trajectory of a moving target, which solves the problem of moving target detection and trajectory correction during vehicle movement. Finally, the feasibility of the proposed method was verified by a series of experiments in autonomous driving environments. The results verify the high recognition accuracy and low positional error of the method.

## I. Introduction

Self-driving technology has advanced rapidly in recent years. However, in complex, cluttered, and unseen environments with high uncertainty, autonomous driving systems frequently make the wrong decisions, so advanced autonomous technologies remain in the laboratory stage [1]. The key to improving system reliability is sensors. Although cameras are the most widely used sensor, they are susceptible to interference from light and dust. In comparison, millimeter-wave (MMW) radar performs robustly in adverse weather and has higher-accuracy range and speed detection [2]. Therefore, MMW radar has been widely used in self-driving systems because of its unique advantages that considerably improve system reliability.

To improve self-driving safety, it is important to predict whether a collision will occur in the future from videos, which involves monitoring and tracking the objects in automotive circumstances. Due to the development of image processing and deep learning technology, tracking algorithms based on cameras have been well-studied[3]; however, radar-based tracking study is limited by several defects. First, the noise and clutter in an environment can lead to false alarm targets and detection errors in the case of a low signal-to-noise ratio (SNR). Second, generally, the radar data only contain the position, velocity, and intensity, which makes the data difficult to classify. Regardless, some research [4-5] achieved simple classification through machine learning methods based on special features such as the doppler pattern, 3D data cube, and statistical features. The plot-level features are scarce, so they are hard to use to support complex intelligent algorithms. To address the problem of tracking moving targets using scarce features, a previous study [6] achieved human-vehicle classification using cluster-level feature vectors, considering the shape parameters, which contain more characteristics than plot-level features. Huang et al. [5] used features in a long time span to distinguish the walking patterns of different people. Their common idea is to aggregate small features into a larger whole to increase the distinguishability of targets. Building on this method, we used clusters as the basic unit of tracks. The typical clustering method is density-based spatial clustering of applications with noise (DBSCAN), but with this method, plots are often mistakenly merged or split apart [4,7]. To effectively exploited plot-level features in the detection, we improve the clustering algorithm by taking distance, velocity, and amplitude into consideration.

The main contribution of this paper is a new data association method in tracking algorithm, which seeks to determine which plots should be used to update which tracks. There are lots of data association method base on radar have emerged [8], such as joint probabilistic data association (JPDA) [8,9], multiple hypotheses tracking (MHT) [8], and the probability hypothesis density (PHD) filter [10]. However, the literature [11] reports that these methods obtain acceptable results in simulation but are flawed in real scenes with noise and scarce prior knowledge. Yang et al. [12] stated that the calculations in these methods are highly complex, limiting their application. To solve that problem, Chen et al. [13] proposed measuring the correlation between radar plots using their feature vectors since the plots between frames have structural similarity. Additionally, it is common to measure the similarity of pixels in computer vision [14]. So, it is possible to classify different clusters consisting of plots in radar graphs. In this paper, our data association algorithm defines a similarity function between clusters to measure the correlation. In every scan, the similarity between two clusters is calculated by the cluster-level features with assigned different weights.

The rest of the paper is organized as follows: In Section II, the system model of the MMW radar is illustrated. In Section III, the clustering algorithm for detected plots is provided, and the proposed target tracking algorithm is illustrated. The experimental result and evaluation are presented in Section IV. This is followed by the conclusions in Section V.

## II. SYSTEM MODEL

Moving target tracking is divided into three steps: target detection, target tracking, and track extraction. In the target detection step, which is a fundamental step, plots are detected and generated the candidate clusters. In the target tracking step, the clusters are assigned to tracks by their similarity. Finally, in the extraction step, the tracks are recognized from moving targets and they are restored to the original tracks by the radar's kinematic parameters.

In the target detection step, the radar estimates the target's range, velocity by 2D-FFT algorithm. Then phase error elimination and beamforming algorithm was used to estimate azimuth. After that, we apply constant false alarm rate (CFAR) to filter out the noise and clutters, and clustering.

In the target tracking step, a weighted feature similarity algorithm in the data association step. After that, we implement a Kalman filter for state prediction and estimation, where the associated cluster revises the latest track prediction to provide an optimal estimate of the target state. The track management system determines which tracks need to be updated or deleted.

Finally, it's the step of moving the target's track extraction. Giving the radar the advantage of detecting moving objects, it can be used to track objects that need the most attention in the driving environment. Therefore, the solution to exclude statics objects from a mobile radar is simply introduced.

## III. PROPOSED ALGORITHM

In this section, the proposed algorithm is presented, including its rationale and implementation.

### A. Clustering

After obtaining plot information from low-level processing, we apply CFAR which can efficiently extract the candidate plots for later clustering. Then, we only keep the maximum amplitude plot in the neighborhood. This idea borrows from maximum pooling in deep learning, and it has the following advantages: 1) extracting the maximum element can retain the main features and eliminate the unimportant ones. 2) making the plots more distinguishable, which is beneficial to the development of the subsequent clustering algorithm. 3) it reduces the point density and simplifies the subsequent data processing of the calculation.

The plots are then divided into several clusters by a clustering algorithm. It is based on DBSCAN, which is widely used in radar signal processing. However, we proposed two improvements. 1) For small obstacles, the plots are scarce and prone to being wrongly recognized as noise by the conventional DBSCAN. Our proposed algorithm assumes that the noise is likely to have a lower amplitude and adds amplitude as a judging condition. 2) It improves the standard for plot clustering by including distance, velocity, and amplitude, which lowers the possibility of wrongly merging different targets. This idea is based on the assumption that the plots from the same target remain close and have similarities in velocity and amplitude. The clustering algorithm is described in Algorithm 1.

**Algorithm 1** clustering

**Input:** $n$-th plot feature vector $p_n = (x_n, y_n, A_n, V_n)$, where the four variables represent the $x,y$ position, amplitude, and velocity, respectively; radius of epsilon neighborhood $\epsilon$, neighborhood point density threshold *MinPts*, amplitude threshold *A_thres*, and velocity threshold *V_thres*.
**Output:** Cluster index $k$, cluster subset $C = \{\Omega_1, \Omega_2, \cdots, \Omega_k\}$

**Variables:** Epsilon neighborhood $N_\epsilon(p_i)$: for $p_i \in P_k$, $N_\epsilon(p_i) = \{p_j \in P_k | dist(p_i, p_j) < \epsilon\}$;
The average amplitude $\bar{A}$ of points in $P_k$, and point set $\Omega$ from the same cluster

1: Initialization: $\Omega = \emptyset, k = 0$, mark all the points as unvisited
2: Noise judgment: $\{p_{noise} | N_\epsilon(p_{noise}) < MinPts \wedge A_{noise} < \bar{A}\}$
3: Find the unvisited non-noise point $p_i$ and its epsilon neighborhood $N_\epsilon(p_i)$
4: Combine $p_j = \{p | p \in N_\epsilon(p_i)\}$ into cluster $p_i$ when $\{p_j | |N_\epsilon(p_j)| \geq MinPts \wedge |V_j - V_i| < V_{thres} \wedge |A_j - A_i| < A\_thres\}$, and mark $p_j$ as visited
5: Return to step 3 until all the eligible points have been considered

The result of the clustering algorithm is the cluster index of plots. In the next part, the proposed tracking algorithm is based on the cluster's features.

### B. Tracking

The proposed tracking algorithm matches the clusters in multiple frames depends on the similarity of the clusters. The similarity is calculated by the cluster's feature vector. In frame $m$ and cluster $k$, the cluster feature vector is defined as

$$F_{m,k} = [P_x, P_y, V_r, S, A, B_{X\max}, B_{X\min}, B_{Y\max}, B_{Y\min}]^T \quad (1)$$

where $P_x, P_y, V_r$ denote the average $x$ and $y$ coordinates value of points and the average radial velocity in cluster $k$, respectively; $A$ is the average amplitude of plots in cluster $k$; $B_{X\max}, B_{X\min}, B_{Y\max}, B_{Y\min}$ are the maximum and minimum $x$- and $y$-coordinate values of the bounding box, respectively; and $S$ is the area of the box.

After obtaining the feature vector, the same cluster between two adjacent frames can be determined by similarity function, which is used to measure the similarity of the feature vectors of two clusters. The similarity contains five parts: distance, velocity, area, overlap, and amplitude, as shown in:

$$Sim = w_1 S_{dis} + w_2 S_{vel} + w_3 S_{area} + w_4 S_{overlap} + w_5 S_{amp} \quad (2)$$

where $\sum_{i=1}^{5} w_i = 1$ are the importance weights between 0 and 1. For example, a greater $w_2$ will be helpful when extracting moving targets. The distance similarity can be defined as

$$S_{dis} = 1 - \frac{dist(P_1, P_2)}{d_{thres}} \quad (3)$$

The distance similarity is determined by Euclidean distance of two clusters' centroids $P_1, P_2$ and $d_{thres}$ is the parameter depends on the radar detection resolution. Analogously, the velocity and area similarity are described as follows

$$S_{vel} = 1 - \frac{|\Delta V_r|}{V_{thres}} \quad (4)$$

$$S_{area} = 1 - \frac{|\Delta Area|}{Area_{thres}} \quad (5)$$

The overlap similarity is defined by the ratio of the intersection and the union of bounding boxes, as shown in (5). It is based on the assumption that the same cluster's boxes in two adjacent frames have the maximum overlapped area. Overlap similarity is 1 if the two bounding boxes coincide; otherwise, it is less than 1.

$$S_{overlap} = \frac{Box_1 \cap Box_2}{Box_1 \cup Box_2} \quad (6)$$

Finally, the amplitude similarity is defined in (7).

$$S_{amp} = 1 - \frac{|\Delta Amp|}{\max(Amp_1, Amp_2)} \quad (7)$$

From the equation above, it can be seen that the similarity is between 0 and 1. The similarity measures the possibility and determines whether a new cluster belongs to the track by considering several factors.

Next, to generate the prediction and estimate the motion of targets, a kinematic model needed to be built. We use a Kalman filter to predict new clusters and provide an optimal estimate when a new cluster is matched, as shown in Algorithm2.

**Algorithm2** Kalman Filter

**Input:** state matrix $A$, new measurement state $z$, frame interval time $\Delta t$, state vector $x = [P_x, P_y, V_x, V_y, B_{X\max}, B_{X\min}, B_{Y\max}, B_{Y\min}]^T$

**Output:** Optimal feature estimation $F_t$

**Variables:** Measure model $H$, state transition model $A$, Kalman gain $K$, process noise covariance $Q$, measurement noise covariance $R$, estimate noise covariance matrix $P$, innovation matrix $y$

------prediction------
1: predict state: $x_{t|t-1} = x_{t-1|t-1} A$

2: predict covariance: $P_{t|t-1} = A P_{t-1|t-1} A^T + Q$

------match------

3: calculate the similarity between the track's prediction feature and clusters in the current frame $t$

4. **if** maximum cluster similarity exceeds the acceptance gate **then**

5. extract the cluster feature and turn it into a measurement state $z$

------update-------

7. innovation: $y_t = Z_t - H x_{t|t-1}$

8. Kalman gain: $K_t = P_{t|t-1} H^T (R + H P_{t|t-1} H^T)^{-1}$

9. Estimate state: $x_{t|t} = x_{t|t-1} + K_t y_t$

10. Update covariance: $P_{t|t} = (I - K_t H) P_{t|t-1}$

11. Predict the cluster feature in the next frame: $x_{t+1} = x_{t|t} A$

12. Transform the state vector $x_{t|t}$ into the feature vector $F_t$, then update the tracks

Note that the prediction and update are separate. In the prediction step, we need to predict the state of the next frame. If the corresponding cluster is found in the next frame by the similarity function, then it will enter the updating step and generate optimal estimation.

Notably, the feature vector contains tangential velocity estimation $V_y$ which is different from the feature vector. It can be derived using tangential displacement, as shown in Fig. 1. The parameters can be expressed as

$$V_t = \frac{r_{k-1} \sin \theta}{\Delta t} \quad (8)$$

$$V_x = V_r \sin \varphi + \xi V_t \cos \varphi \quad (9)$$

$$V_y = V_r \cos \varphi + \xi V_t \sin \varphi \quad (10)$$

A factor $\xi \in [0,1]$ is set to reduce the influence of position error caused by azimuth inaccuracy on the prediction result, which will achieve better experimental results.

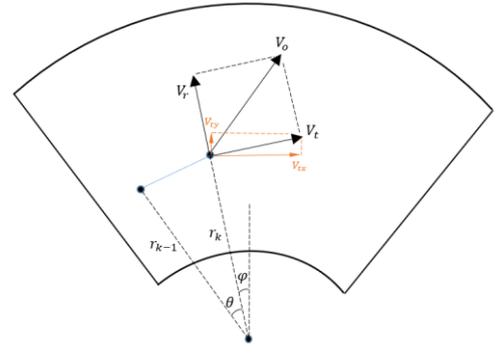

Fig. 1. Tangential velocity estimation model, where $V_0$ denotes the target velocity in the current frame, which is decomposed into radial velocity $V_r$ and tangential velocity $V_t$. The two points are the cluster centroids of the current frame and the previous frame.

To differentiate the static target from the moving one requires the motion parameters of the radar. Suppose the velocity components of radar are $V_x$ and $V_y$, the azimuth is $\alpha$, as shown in Fig.2. The radial velocity of static targets can be derived as

$$V_{static} = V_{rx} + V_{ry} = V_x \sin \alpha + V_y \cos \alpha \quad (11)$$

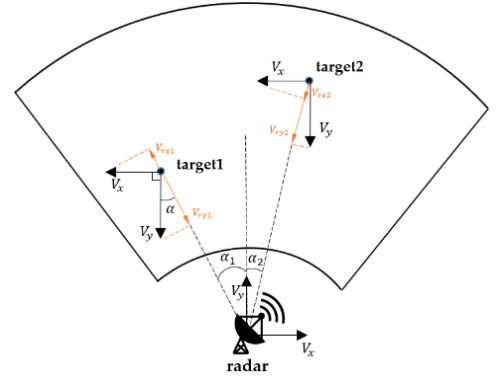

Fig. 2. Decomposition of the relative velocity of static targets.

$V_{static}$ is only affected by the azimuth when the radar velocity is set. Here, the compensation involves subtracting $V_{static}$ from the detected radial velocity.

$$|V_r - V_{static}| > \delta_v \quad (12)$$

Ideally, the velocity of static objects will be 0 after compensation. However, in practice, because of noise and error, the final error is acceptable within a certain threshold $\delta_v$, whose value depends on the real circumstance; for example, on a road with high-speed cars, a larger $\delta_v$ would perform better; In a street with bicycles and pedestrians, $\delta_v$ should be set lower.

## IV. PERFORMANCE EVALUATION

In this section, we evaluate the performance of the method through a series of experiments. To test the object detection performance and target tracking algorithm of our method, three experiments were conducted.

The experiments are based on Texas Instrument's (TI) automotive MMW radar products. We chose AWR1642 as the radar test platform, having four receiving channels and two transmitting channels, and 4 GHz available bandwidth from 76 to 81 GHz. Then, we chose DCA100 EVM as the data capture board for streaming the ADC raw data from the AWR1642 to the computer through an ethernet cable.

Previous studies [15,16] introduced a rich set of metrics for tracking system evaluation of camera systems, which can also be used in radar tracking. Given these metrics, we evaluated the performance of our method from three aspects: centroid matching error (CME), bounding box overlap rate (BBOR), and F1-score. The CME and BBOR are defined as

$$CME = \frac{1}{N}\sum_{i=1}^{N} dist(GT_i, DT_i) \quad (11)$$

$$BBOR = \frac{1}{N}\sum_{i=1}^{N} \frac{GT_i \cap DT_i}{GT_i \cup DT_i} \quad (12)$$

where $N$, $GT$, and $DT$ represent the length of track, ground truth, and detection, respectively. CME and BBOR represent the centroid position error, whereas BBOR represents the shape error.

### A. Static detection performance

In the first experimental scenario, multiple targets were detected by a static radar to produce clustering results. We chose a wide road and placed the targets in a fixed position, and the size of the targets was measured to calculate the BBOR. The experimental scene are shown in Fig. 3.

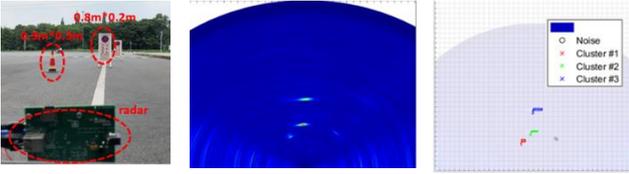

Fig. 3.   Target detection scene and the result of the algorithm

Used dataset from three detections, we got an average CME of 0.193m, and an average BBOR of 78.8%. Considering the error is limited by the range resolution (0.12m), the CME of the stationary target satisfies the precision requirement. The high overlap degree of the bounding box is due to the small change in the object echo, which makes the size of the cluster box relatively fixed. To summarize, the multi-target detection algorithm has high accuracy and meets the basic experimental requirements.

### B. Tracking Performance

The second experiment was for the tracking algorithm; we used a static radar to detect moving targets including pedestrians, bicycles, and sedans, which are the most common targets in automotive scenarios. The target moved in both the radial and tangential directions. Three non-target clutter items were placed in the environment. The reason for the tangential direction was to evaluate the KF prediction.

The problem of ground truth measurement while tracking must be noted. The ground truth can be obtained by a more precise sensor such as a camera, but this experiences many problems (e.g., time synchronization). After comprehensive consideration of accuracy and simplicity, we chose to drive the radar vehicle with a constant velocity on a straight line with fixed starting and ending points. So, we assumed that the actual centroid of the target was equidistant on the line segments. Then, the measured size of the targets was assumed to be the actual bounding box that was used to calculate the BBOR.

Five sets of data were collected in each direction of motion, and the performance was evaluated by three indicators (CME, BBOR, and F1-score). The test scenario is shown in Fig. 4.

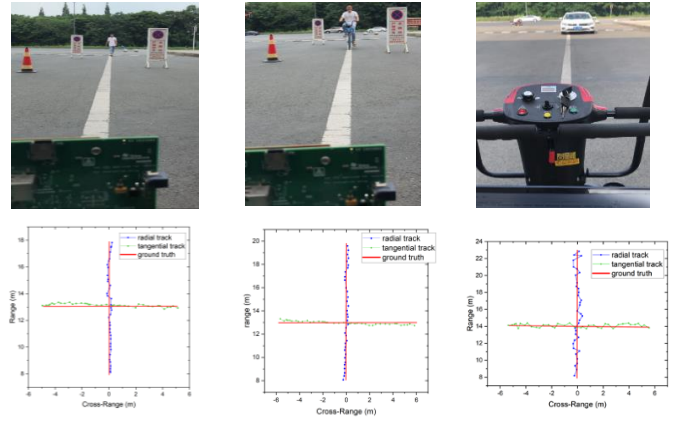

Fig. 4.   Target detection scene and the result of the algorithm for different targets.

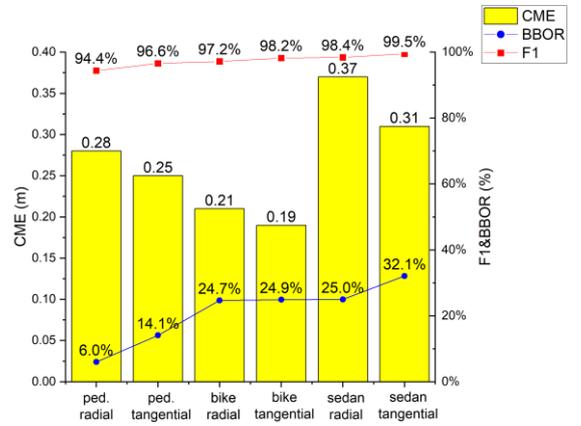

Fig. 5.   Performance of tracking algorithm

Several patterns are shown in Fig. 5. For the same kind of target, the tangential track has a lower CME, and higher BBOR and F1-score, which indicate higher accuracy. This result may be due to the detection range, as a radial track with a further start point may contain more noise and error.

When comparing the different targets, we found three characteristics. First, the order of the F1-score(tracking correctness) is sedan, bicycle, and pedestrian. The tracking correctness mainly depends on target size and material, a target of large size and made of metal are prone to have an intense electromagnetic wave echo, which makes them easier to observe by radar. Second, the sedan's BBOR is the highest, and the pedestrian's is the lowest. Apparently, a large target is favorable for the overlapping of the adjacent bounding box. Third, the bicycle received the lowest CME, for two potential reasons. The first reason is positional error, due to the bicycles move more straightly than people walking. The second reason is the size of the bounding box: a larger box leads to a larger variation in the centroid. Due to the detected plots of sedan clusters being distributed unevenly in the bounding box, the centroids fluctuate and the CME increases.

### C. Trajectory Correction

The third experiment was used to evaluate the trajectory correction performance. The radar was mounted on a vehicle.

The vehicle and target moved straight at constant velocity simultaneously to follow a fixed line on the ground.

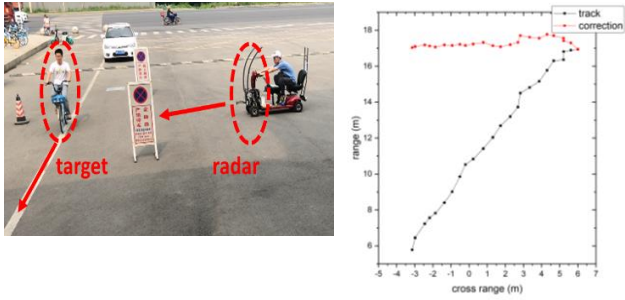

Fig. 6. Example scenario of the trajectory correction process.

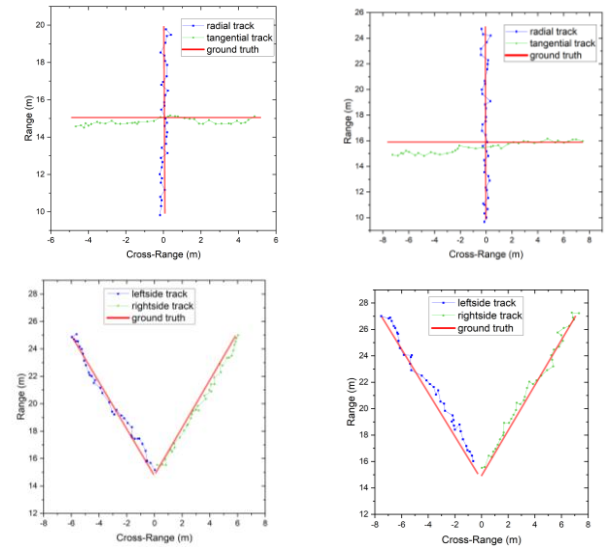

Fig. 7. Tracking algorithm result of different targets and directions.

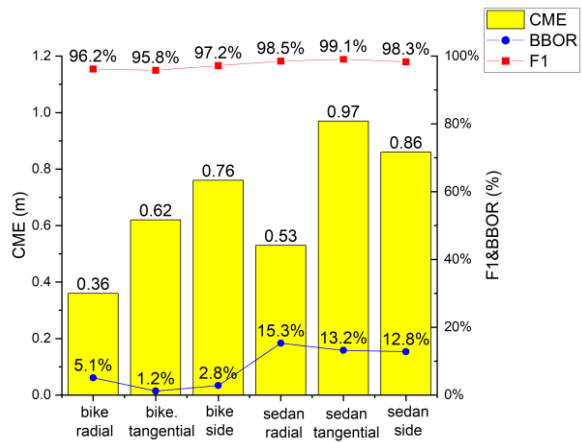

Fig. 8. Target detection scene and the result of the algorithm

According to Fig. 7, the tracks showed a large deviation and lower positional accuracy from the ground truth compared to the second experiment. In combination with the analysis of the actual scene, the errors of the radar motion and the target motion were superposed so that it was difficult to restore the trajectory to a straight line after correction. The velocity variation in the radar vehicle also led to increased tangential track error.

However, there was no significant degradation in the F1-score, meaning that the target could still be effectively distinguished from the environment, indicating that the target detection and tracking algorithms perform well. A high precision sensor on the radar vehicle is the key to improving the performance in the third experiment.

V. CONCLUSION

In this study, an algorithm for moving target detection and tracking was constructed, and the performance of the algorithm was tested with an automotive radar. In target detection, we used the conventional method including the 2D-FFT algorithm and the capon beamforming algorithm, then the main target plots were extracted by CFAR. A improve clustering method was applied to the plots. In moving target tracking, the clusters were assigned to different tracks by the feature similarity matching method. Then, the trajectory of the moving target was identified and restored. The experimental results in real road scenes showed that the proposed algorithm performs well with high accuracy and robustness.